# Efficient Road Lane Marking Detection with Deep Learning

Ping-Rong Chen*, Shao-Yuan Lo*, Hsueh-Ming Hang
National Chiao Tung University
a6561814@gmail.com, sylo95.eecs02@g2.nctu.edu.tw,
hmhang@nctu.edu.tw

Sheng-Wei Chan, Jing-Jhih Lin
Industrial Technology Research Institute
{ShengWeiChan, jeromelin}@itri.org.tw

*Abstract*—Lane mark detection is an important element in the road scene analysis for Advanced Driver Assistant System (ADAS). Limited by the onboard computing power, it is still a challenge to reduce system complexity and maintain high accuracy at the same time. In this paper, we propose a Lane Marking Detector (LMD) using a deep convolutional neural network to extract robust lane marking features. To improve its performance with a target of lower complexity, the dilated convolution is adopted. A shallower and thinner structure is designed to decrease the computational cost. Moreover, we also design post-processing algorithms to construct 3$^{rd}$-order polynomial models to fit into the curved lanes. Our system shows promising results on the captured road scenes.

## I. INTRODUCTION

In general, lane detection algorithms include the following steps: (1) lane marking generation, (2) lane marking grouping, (3) lane model fitting, and (4) temporal tracking. Extracting the correct lane is critical for successful lane-mark generation. Many conventional approaches detect the lane using the information of edge [4, 5], color, intensity and shape. In addition, lane detection can be viewed as an image segmentation problem [6]. However, most methods are sensitive to illumination changes, weather condition and noises; and thus many traditional lane detection systems fail when the external environment has significant variation.

In recent years, the Deep Convolutional Neural Network (DCNN) based methods have been proposed and they outperform the traditional approaches on many applications. It also demonstrates a huge success on image semantic segmentation; we thus use this technique to extract stable lane features. Fig. 1 shows the flowchart of a lane marking detector (LMD) system. A CNN-based method is used to produce lane-marks in the first step, but the other two steps still adopt the traditional approaches. For a typical CNN, down-sampling is employed to enable a deeper architecture and also to enlarge the receptive field to capture large-scale objects in images. However, this operation usually reduces the detailed spatial information, which is very important for a semantic segmentation task. Therefore, several network architectures have been proposed to solve the problem.

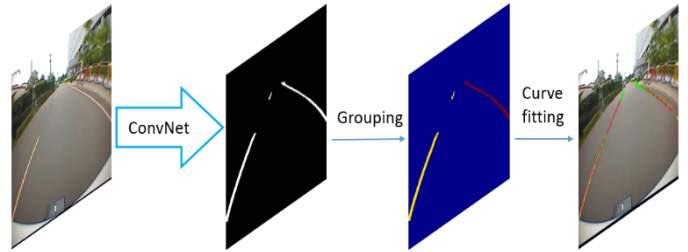

Figure 1. Flowchart of the proposed LMD system.

In general, two methods are used to recover or retain detailed information. The first one is using an encoder-decoder architecture. The encoder is similar to many classification networks, such as VGG [7] and ResNet [8]. The decoder consists of consecutive up-sampling operations in order to reconstruct the same resolution as the input image. Deconvolution, a learnable up-sampling layer, is the most common approach to up-sample the feature maps, such as DeconvNet [9], FCN [10] and U-Net [11]. After up-sampling, FCN [10] and U-Net [11] use the feature maps directly coming from the encoder to recover more details. SegNet [1] applied another up-sampling method by transferring the max-pooling indices from encoder to decoder. It tends to be more efficient in term of memory usage because of storing fewer indices.

The other approach is using the dilated convolution [2, 3]. This method removes some down-sampling layers so that the feature maps can maintain spatial resolution and thus retain details. Nevertheless, removing down-sampling is not favorable to have large receptive fields. Thus, the dilated convolution is employed to enlarge the receptive fields at different rates. Since lane detection is a real-time application, reducing computation is of high priority. In this work, we combine the advantages of the above two approaches and modify the system to achieve low complexity and maintain similar accuracy.

## II. NETWORK ARCHITECTURE

The architecture of LMD network is shown in Fig. 2. It is an encoder-decoder architecture that follows the structure similar to U-Net [1] and SegNet [2]. The encoder is a variant of VGG16 [7]. It consists of 14 3x3 convolutional layers; the first 13 convolutional layers correspond to the convolutional layers of VGG16, and the last one is inserted for matching the number of feature channels of the decoder input. Each convolutional layers are followed by a batch normalization layer [12] and a rectified linear unit (ReLU).

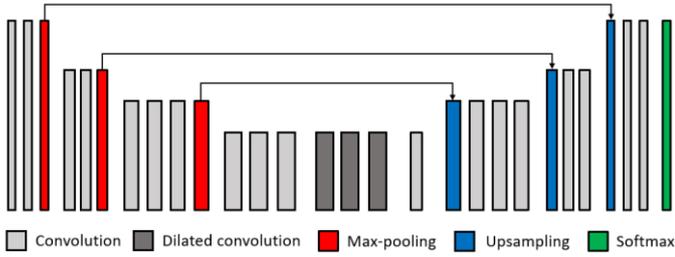

Figure 2. Architecture of the proposed LMD network.

We discard the fully-connected layers of VGG16 network to decrease the number of parameters for speed consideration and to maintain the feature map resolution for accurate localization purpose [3]. Since the VGG16 network is designed for image classification, it consists of many max-pooling layers for downsampling. However, the downsampling operation loses a significant amount of spatial information, which is critical for semantic segmentation. In order to solve the spatial localization problem, we only retain the first three 2x2 max-pooling layers with stride 2 in VGG16 that located right after the 2nd, 4th and 7th convolutional layer respectively. So, the resolution of the feature maps at the end of the encoder network is increased by a factor 4. This modification enables our network to capture small classes and boundary details.

Because the feature maps are enlarged after the 10th convolutional layers, the convolutional kernels of the following layers have to be enlarged by the corresponding factor to keep the same receptive field of the network. We implement this efficiently by employing the dilated convolution [2, 3]. Dilated convolution is dilating the convolutional filters by expanding its size and filling the empty positions with zeros, which can expand its receptive field without any additional parameters and computational cost. Thus, we set the 11th to 13th convolutional layers with dilation 2.

The decoder in SegNet, which is responsible for recovering the image resolution, is an exact mirror of the encoder. In contrast, benefited by using larger feature maps, the decoder of LMD can be considerably simplified, which consists of only 7 convolutional layers and 2 max-unpooling layers. We use thinner convolutional layers in the decoder for speeding up without sacrificing the accuracy. The indices of the max locations computed by the max-pooling units in the encoder are stored and passed to the corresponding max-unpooling layers in the decoder for upsampling. Finally, the softmax classifier is inserted after the decoder for pixel-wise classification.

### III. CLASS WEIGHTING ANALYSIS ON LANES

We use CamVid [14] dataset (the details are in Sec. V) and take SegNet [1] to investigate how class weighting affects the output results on the lane class. First, we apply the median frequency balancing scheme to compute the class weights for all the classes in this dataset. Then, we adjust the class weight of lane by multiplying the factor of 0.6, 2 and 5, respectively.

Table I shows the results. The class accuracy is positively correlated with the class weight, while the IoU is negatively correlated with the class weight. That is, the larger the class weight is, the wider area the lane class will be. These results indicate that the class accuracy metric does not include the false alarm case. Thus, a thicker lane leads to better class accuracy. In contrast, since IoU is punished for false alarms severely, the thinner lanes receive better points. The above analyses show that by adjusting the class weight, we can change the thickness of the segmented lane class according to the requirement of the lane mark post-processing operations (e.g., grouping, curve fitting).

TABLE I. THE IMPACT OF WEIGHT ADJUSTMENT

| The class weight of lane | Class accuracy | IoU |
| --- | --- | --- |
| Balanced * 0.6 | 80.7 | 53.6 |
| Balanced | 83.9 | 52.6 |
| Balanced * 2 | 82.9 | 51.3 |
| Balanced * 5 | 88.7 | 47.3 |

### IV. POST-PROCESSING SCHEME

A conventional lane detection algorithm can be divided into three steps: Lane marking generation, Lane Grouping and Lane Model Fitting. As introduced in the previous sections, we generate lane-marks by CNN semantic segmentation. We train LMD with a three-class dataset, including lane, road and the others. After lane detection by LMD, lanes are extracted as binary images. In this section, we explain the next two steps.

#### A. Lane Grouping

The concept of lane grouping consists of two steps. The first step is to cluster neighboring pixels belonging to the same lane-segment to form a supermarking [13]. Different from the approach used in [13], the connected component labeling (CCL) technique is used to detect the connected regions and assign one label value to each region. One region is described only by one label value instead of by lots of pixels, thereby reducing the complexity problem. The second step is to connect supermarkings, which are on the same lane marking. It is very important to design measurement functions properly to calculate the cost of connecting supermarkings. Inspired by [13], our measurement functions are defined by Fig. 3. However, the distributions of these two costs may be different, and thus it results in a hard way to combine them. To solve the problem, we normalize the distribution of directionality into the distribution of geometric distance, as (1) shows.

$$\text{dist}'_{ri} = \frac{\text{dist}_{ri} - \text{mean}_{dir}}{\sqrt{\text{var}_{dir}}} \times \sqrt{\text{var}_{geo}} + \text{mean}_{geo} \quad i=1,2,3,4 \quad (1)$$

$$\text{cost} = \frac{1}{4} \sum_{i=1}^{4} \left( \frac{\text{dist}_{gi} + \text{dist}'_{ri}}{2} \right) \quad (2)$$

where $\text{mean}_{dir}$ and $\text{mean}_{geo}$ are the mean of the four $\text{dist}_{ri}$ and $\text{dist}_{gi}$ in Fig 3. And $\text{var}_{dir}$ and $\text{var}_{geo}$ are the variance of the four $\text{dist}_{ri}$ and $\text{dist}_{gi}$, i=1~4. After (1), the distributions of these two costs are identical so that we can combine these two costs by a simple average operation, as written in (2). If the cost of connecting two supermarkings is sufficiently small, they belong to the same lane-marking and thus should be connected to each other. Hence, they are now assigned to the identical label value.

#### B. Lane Model Fitting

Lane model fitting is used to represent both straight lanes and curve lanes. So, we adopt the 3$^{rd}$ order polynomial for the lane model, which can describe a high curvature lane,

$$y = ax^3 + bx^2 + cx + d \quad (3)$$

where *x* and *y* are coordinates, *a, b, c* and *d* are the coefficients determined by a series of data points. After grouping, it becomes easier to fit curves to the properly assigned label values. However, due to an amount of lane features detected by segmentation model, curve fitting is a tedious work. To solve this problem, we divide the image into several blocks along its vertical axis, then we compute the mean position of lane features in every block. By doing so, a lane is fitted by very few candidates instead of a number of lane feature points. After the lane model fitting step, the coefficients of a polynomial of degree three are determined by fitting. A 3$^{rd}$–order model is adopted to match the curved lanes with rather large curvature.

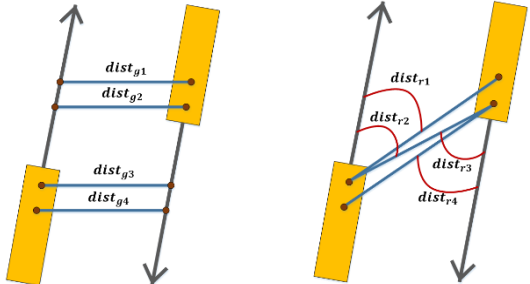

Figure 3. Specifications of supermarking connecting cost.

## V. EXPERIMENTS

We conduct a series of experiments for validating the performance of LMD. The details are given below.

**Dataset:** We include the CamVid [14] road scene dataset to the performance benchmark on LMD. This dataset contains 367 training and 233 testing RGB images at 360x480 resolution. There are 11 classes, such as road, cars, traffic signs, sky, etc. The undefined class is ignored during training. We compare our results with the state-of-the-art methods based on these classes. Moreover, we also perform experiments for 12 classes, in which the lane class is added.

**Setup:** All the experiments are conducted on the Caffe [15] platform with cuDNN v5.1 back-end and running on a single GTX 1080 GPU. The popular stochastic gradient decent (SGD) algorithm is adopted to train LMD. We use the ImageNet [16] pre-trained model of VGG16 [7] as the initial values of our encoder. Training is conducted with a weight decay of 5e-4, momentum of 0.9, batch size of 3, and the initial learning rate is set to 0.01 and is decreased by a factor of 10 after each 5,000 training iterations, and totally there are 20,000 iterations.

Since the number of pixels varies significantly among the classes in the training set, we employ a class balancing technique to differently weight the loss according to class frequencies. We use the median frequency balancing scheme [17] that specifies the class weight of each class as $w_c$ = median($p$) / $p_c$, where $p_c$ is the number of pixels of class $c$ divided by the total number of pixels in images where $c$ is present, and median($p$) is the median of these frequencies.

TABLE II. COMPARASIONS WITH FORWARD TIME AND MODEL SIZE

| Network | Inference time (ms) | Frames per sec. (fps) | Model size (MB) |
|---|---|---|---|
| SegNet [1] | 35.5 | 28.1 | 117 |
| LMD (ours) | 29.1 | 34.4 | 66 |

### A. Performance Analysis

Table II compares the forward inference time (fwt) and the number of frames per second (fps) associated with a single input image of 360x480 resolution. These are computed by the Caffe time command. The model size is referred to the Caffe model size on the hard disk. LMD is faster than SegNet by 22.4% with a significantly smaller model size. These results show that LMD is more suitable for real-time applications and also more portable.

### B. Accuracy Evaluation

We evaluate the performance of LMD on the CamVid dataset by class accuracy and intersection-over-union (IoU) metrics. Table III displays our results and the comparisons with two state-of-the-art segmentation models: SegNet [1] and ENet [18], which are also designed for faster inference speed, fewer parameters and requiring less memory.

LMD significantly outperforms the other models on both class average accuracy and mean IoU (mIoU) metrics. Particularly, it achieves considerable improvement on all the challenging classes (e.g., traffic signs, pedestrian, bicyclist), which contain less training samples or complicated shapes. The segmentation results are provided in Fig. 4. Furthermore, we add the lane as the 12th class for training and testing because it is indispensable to autonomous driving applications. With this additional class, LMD still surpasses the other models. Finally, we combine CamVid [14], KITTI [19] and SYNTHIA [20] to form an ensemble of 4,004 images for pre-training; it achieves a better performance at a 79.6% class average accuracy and 65.2% mIoU. Apparently, the method we applied that enlarges the feature maps by using dilated convolution is effective.

### C. Results after Post-Processing

Fig. 5 demonstrates the experimental results of all the steps in our lane detection algorithm. Fig. 5(a) is an input image, and Fig. 5(b) shows the segmentation result, which is trained on 3 classes: road, lane and others. Fig. 5(c)-(d) are the results in the post-processing stage, including grouping and curve fitting. Each color represents a road-marking in Fig. 5 (c), and the green points in Fig. 5(d) are candidates introduced in the section of Lane Model Fitting. Fig. 5(e)-(h) show the processed results of 4 road scenes, and our algorithm is able to achieve high quality in all cases. It is sufficiently stable to get over the complicated environment even when there're lots of distraction signs on road.

## VI. CONCLUSIONS

In this paper, we employ encoder-decoder architecture, dilated convolution and fine-tuned modifications to develop a modified CNN for road lane detection, called Lane Mark Detector (LMD). We improve the accuracy to achieve 65.2% mIoU on the CamVid dataset, and we also improve the testing speed to 34.4 fps. For ADAS applications, we combine the idea in [13] to develop a simple post-processing algorithm, and to construct an accurate 3$^{rd}$-order lane model. The experimental results indicate that our model is stable and is able of tolerating many variations on road scenes.

ACKNOWLEDGMENT

This work was supported in part by the Mechanical and Mechatronics Systems Research Lab., ITRI, under Grant 3000518795.

TABLE III. COMPARISONS WITH CLASS ACCURACY AND mIoU ACCURACY

| Network | Buil. | Tree | Sky | Car | Sign | Road | Pede. | Fenc. | Pole | Side. | Bike | Lane | Class avg. | mIoU |
|---|---|---|---|---|---|---|---|---|---|---|---|---|---|---|
| SegNet [1] | 88.8 | **87.3** | 92.4 | 82.1 | 20.5 | **97.2** | 57.1 | 49.3 | 27.5 | 84.4 | 30.7 | - | 65.2 | 55.6 |
| ENet [18] | 74.7 | 77.8 | **95.1** | 82.4 | 51.0 | 95.1 | 67.2 | 51.7 | 35.4 | 86.7 | 34.1 | - | 68.3 | 51.3 |
| LMD-11 | **89.2** | 86.4 | 93.7 | **83.8** | **58.1** | 95.4 | **79.3** | **52.7** | **48.6** | **90.5** | **61.6** | - | **76.3** | **63.5** |
| LMD-12 | 88.1 | 86.8 | 94.0 | 84.3 | 55.4 | 90.1 | 80.1 | 51.9 | 48.4 | 92.3 | 64.7 | 83.9 | 76.7 | 62.2 |
| LMD-12 (pre-tr.) | 89.3 | 87.9 | 94.1 | 87.0 | 63.7 | 91.2 | 86.0 | 55.2 | 54.8 | 93.9 | 67.0 | 85.4 | 79.6 | 65.2 |

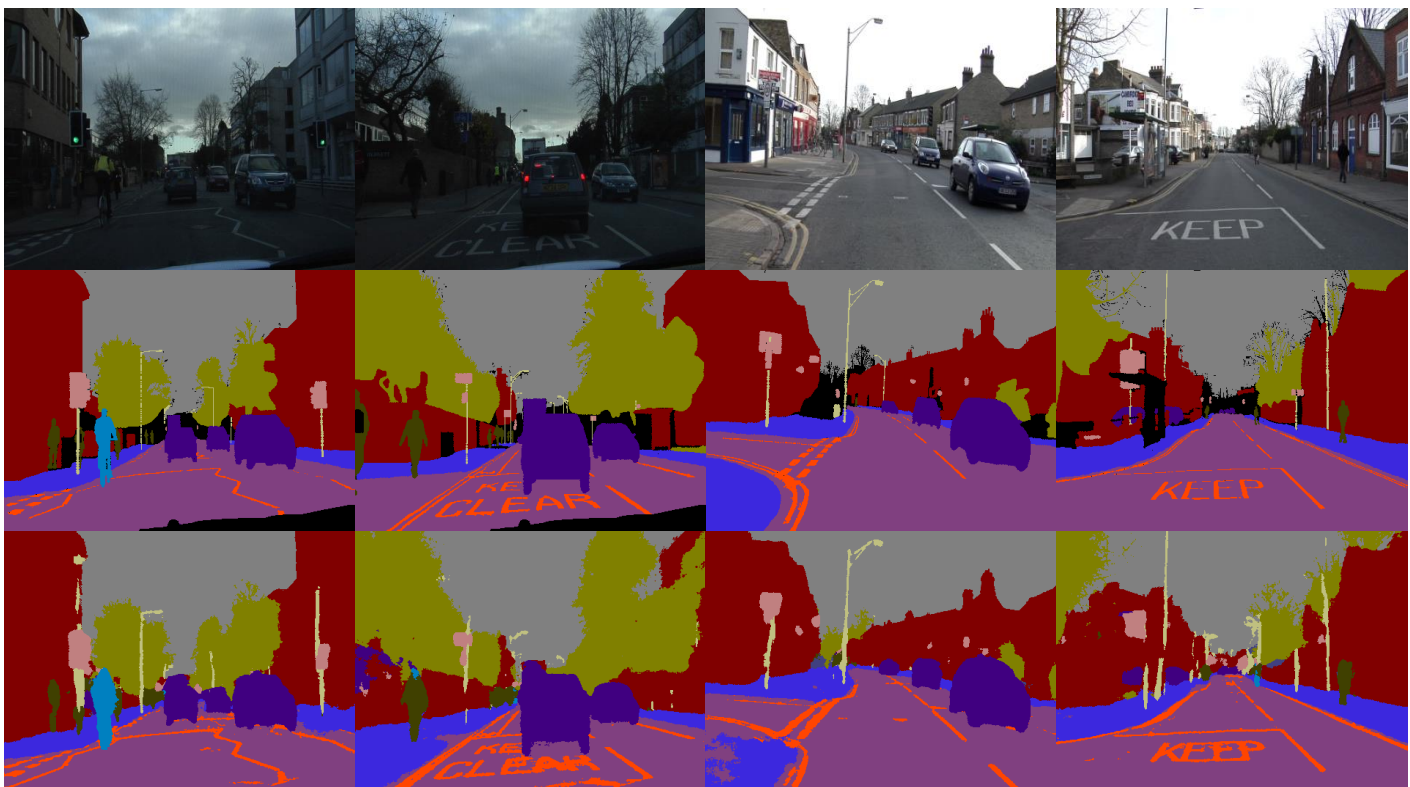

Figure 4. Results of segmentation produced by LMD. From top to bottom: (a) Input image, (b) Ground truth, (c) LMD output.

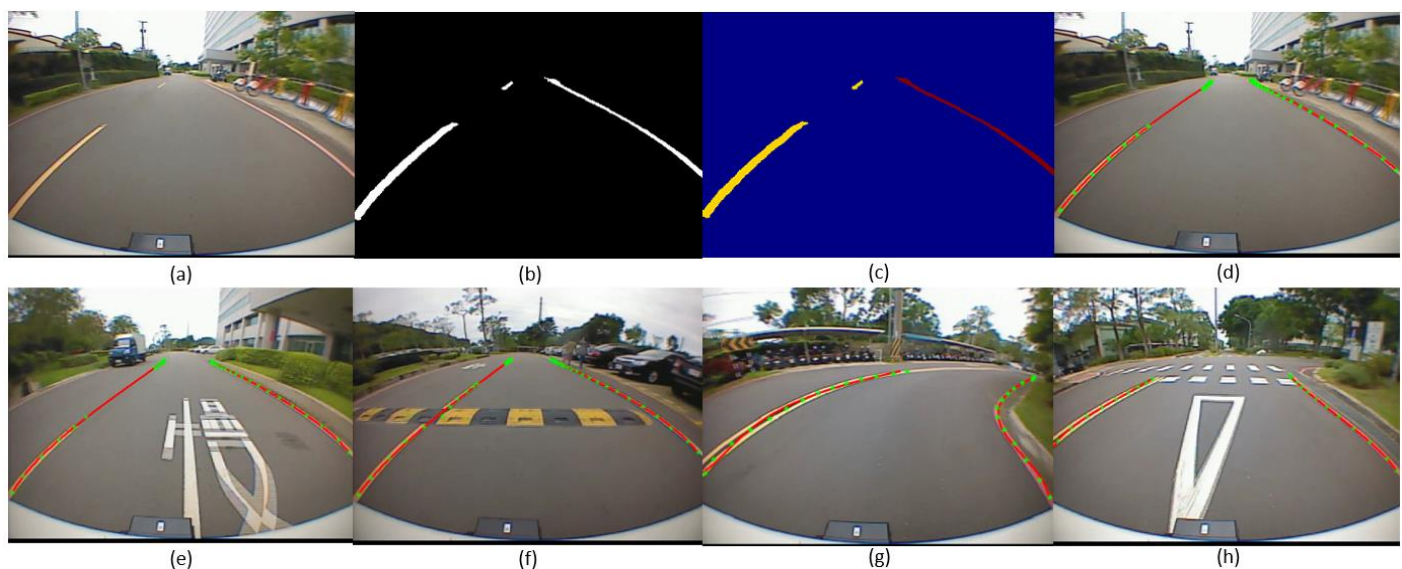

Figure 5. Samples of lane detection results: (a) captured image, (b) detected lane class, (c)-(d) grouping and curve fitting, (e)-(h) final results of 4 test images.